%% file: main.tex
\documentclass[sts]{imsart}

\RequirePackage{amsthm,amsmath,amsfonts,amssymb,graphicx}
\RequirePackage[numbers,sort&compress]{natbib}
\RequirePackage[colorlinks,citecolor=blue,urlcolor=blue,backref=page]{hyperref}

\usepackage{
    amsfonts,
    anyfontsize,
    bbm,
    booktabs,
    caption,
    enumitem,
    graphicx,
    lipsum,
    mathtools,
    microtype,
    nicefrac,
    paralist,
    pdflscape,
    pifont,
    subcaption,
    tikz,
    url,
    wrapfig,
    xargs,
    xcolor,
}
\usepackage[ruled,vlined]{algorithm2e}
\usepackage[capitalise]{cleveref}
\usepackage[acronym,nowarn,section,nogroupskip,nonumberlist]{glossaries}

\startlocaldefs

\theoremstyle{plain}

\theoremstyle{remark}

\endlocaldefs

\usetikzlibrary{backgrounds}
\glsdisablehyper{}

\newacronym{ACE}{ACE}{adaptive contrastive estimation}
\newacronym{BA}{BA}{Barber-Agakov}
\newacronym{BO}{BO}{Bayesian optimization}
\newacronym{EIG}{EIG}{expected information gain}
\newacronym{KL}{kl}{Kullback-Leibler}
\newacronym{MI}{MI}{mutual information}
\newacronym{MC}{MC}{Monte Carlo}
\newacronym{NCE}{NCE}{noise contrastive estimation}
\newacronym{NMC}{NMC}{nested Monte Carlo}
\newacronym{OED}{BED}{Bayesian experimental design}
\newacronym{RMSE}{RMSE}{root mean squared error}
\newacronym{SGD}{sgd}{stochastic gradient descent}

\DeclareMathOperator{\E}{{}\mathbb{E}}

\newcommand{\designnet}{\pi_\phi}

\newcommand{\latent}{\theta}

\newcommand{\prior}{p(\latent)}

\newcommand{\obslik}{p(y|\latent, \xi)}

\newcommand{\obsmarg}{p(y|\xi)}
\newcommand{\obspost}{p(\latent|y, \xi)}

\newcommand{\iid}{\overset{\scriptstyle{\text{i.i.d.}}}{\sim}}

\newcommandx{\histmarg}[1][1=\designnet]{p(h_T| #1)}
\newcommandx{\histlik}[1][1=\designnet]{p(h_T|\latent, #1)}
\newcommandx{\histliki}[2][2=\designnet]{p(h_T|\latent_{#1}, #2)}
\newcommandx{\munmc}[2][1=N,2=M]{\ensuremath{\hat{\mu}_{#1,#2}}}
\newcommandx{\munmcq}[2][1=N,2=M]{\ensuremath{\hat{\mu}_{#1,#2,q}}}

\renewcommand{\to}{\ensuremath{\rightarrow}}

\begin{document}

\begin{frontmatter}

\title{Modern Bayesian Experimental Design}
\runtitle{Modern Bayesian Experimental Design}

\begin{aug}

\author[A]{\fnms{Tom}~\snm{Rainforth}\ead[label=e1]{rainforth@stats.ox.ac.uk}},
\author[B]{\fnms{Adam}~\snm{Foster}\ead[label=e2]{adam.e.foster@microsoft.com}},
\author[C]{\fnms{Desi R}~\snm{Ivanova}\ead[label=e3]{desi.ivanova@stats.ox.ac.uk}},
\and
\author[D]{\fnms{Freddie}~\snm{Bickford Smith}\ead[label=e4]{freddie@robots.ox.ac.uk}}

\address[A]{Tom Rainforth is a Senior Researcher, Department of Statistics,
    University of Oxford\printead[presep={\ }]{e1}.}
\address[B]{Adam Foster is a Senior Researcher, Microsoft Research AI4Science \printead[presep={\ }]{e2}.}
\address[C]{Desi R Ivanova and Freddie Bickford Smith are DPhil Students, University of Oxford \printead[presep={\ }]{e3,e4}.}

\end{aug}

\begin{abstract}
Bayesian experimental design (BED) provides a powerful and general framework for optimizing the design of experiments.
However, its deployment often poses substantial computational challenges that can undermine its practical use.
In this review, we outline how recent advances have transformed our ability to overcome these challenges and thus utilize BED effectively, before discussing some key areas for future development in the field.
\end{abstract}


\end{frontmatter}

\section{Introduction}
\label{sec:introduction}

From clinical trials~\citep{cheng2005bayesian} and market research~\citep{kuhfeld1994efficient}, to particle physics simulations~\citep{melendez2021designing} and drug discovery~\citep{lyu2019ultra}, the design of experiments is a fundamental challenge throughout science and industry.
Better experiments lead to better data, while good quality data is a cornerstone of much of the modern world.
Despite this, quantitative methods for optimizing data acquisition have been far less explored in the statistics and machine learning literatures than how to utilize data once we have it.

At a high level, we want to choose designs that maximize the amount of \emph{information} gathered from our experiments.
This desire can be formalized using \emph{Bayesian experimental design} (BED)~\citep{chaloner1995,lindley1956,lindley1972,sebastiani2000maximum,ryan2016review,cavagnaro2010adaptive}, which provides a powerful model-based framework for choosing designs `optimally' using information-theoretic principles.

BED can be applied whenever we can construct an appropriate model, or simulator, for our experiment, with success stories in a diverse set of fields, such as psychology~\citep{myung2013,vincent2017darc,watson2017quest+}, physics~\citep{dushenko2020sequential,loredo2004bayesian}, bioinformatics~\citep{vanlier2012}, neuroscience~\citep{shababo2013bayesian}, engineering~\citep{papadimitriou2004optimal}, and machine learning~\citep{houlsby2011bayesian,gal2017deep}.
It can be particularly useful when choosing designs manually is difficult.
For example, it may not be intuitively clear what designs will work well, we might want to leverage existing data, or we might need to automate the running of our experiment(s).

BED can be especially effective in \emph{adaptive} design contexts, wherein we have a series of design decisions and can utilize previous outcomes at each step.
Here it provides a natural mechanism for optimizing and automating the data acquisition process, utilizing information gathered during the experiment to dynamically make design decisions on the fly~\citep{mackay1992information,cavagnaro2010adaptive,foster2021dad,huan2016sequential}.
This adaptive, or \emph{sequential}, BED approach has appeared under a variety of names in the literature, such as \emph{Bayesian adaptive design}~\citep{cheng2005bayesian,zhou2008bayesian,vincent2017darc} and \emph{adaptive design optimization}~\citep{cavagnaro2010adaptive,myung2013}, often in the form of special cases, such \emph{Bayesian active learning}~\citep{mackay1992information,houlsby2011bayesian} and \emph{Bayesian optimization}~\citep{hennig2012entropy,hernandez2016predictive,shahriari2015taking,wang2017max}.

While, in principle, there is huge potential for using BED to intelligently acquire data in almost every quantitative field, its uptake to date has arguably been comparatively modest.
Several historical factors have contributed to this~\citep{robertson2023response,ruberg2023application}, such as insufficient expertise in potential application fields~\citep{medical2022not,pallmann2018adaptive}, a past dearth of appropriate regulatory guidance (e.g.~US FDA guidance for adaptive clinical trials was issued in 2019~\citep{usfda2019adaptive}),
and philosophical battles over the use of Bayesian statistics~\citep{gelman2008objections}.
However, there has also been a huge technical barrier to its use as well:
it tends to suffer from some crippling computational bottlenecks that have often prevented its practical deployment~\citep{rainforth2018nesting,zheng2018robust,beck2018fast}, especially in adaptive contexts.

This, though, is an exciting time for BED.
In recent years a number of important developments have managed to systematically tear down some of the key computational bottlenecks, revolutionizing what is potentially achievable with it.
The aim of this review paper is to provide an overview of these developments---outlining how the field has transformed since the publication of previous reviews~\citep{chaloner1995,ryan2016review}---and a guide as to how BED can now be effectively deployed in practice using modern techniques.
We will further provide insights and suggestions on important areas for future development.

\section{Information-Theoretic Design}
\label{sec:background}

Consider an arbitrary experiment with some controllable aspects, $\xi$, known as its \emph{design}.
The choice of this design will influence the experiment's outcome, $y$.
This outcome should be thought of as a random variable before the experiment is carried out, as there would be no need to run the experiment if we could already predict its outcome exactly and we still need to account for our own uncertainty, even if the true underlying process is not stochastic.

The data gathered from the experiment is now uniquely defined by the pair $(\xi,y)$, assuming, by convention, that any auxiliary recordings, such as uncontrollable aspects of the experimental setup, are incorporated into $y$.
The vast majority of experimental design problems can thus be formalized as optimizing the design to maximize some (potentially abstract) expected utility of the acquired data, where this expectation is over the possible outcomes.

The focus of this paper will be on methods that measure the utility of the acquired data through the amount of \emph{information} it provides.
To formulate this mathematically, we need some quantity that we will aim to gather information about; we will denote this as $\theta$.
Here $\theta$ might be some set of real-world quantities, a collection of parameters in a model, or represent something more abstract, such as future model predictions.

\subsection{Bayesian Experimental Design}

In BED, we express our initial beliefs about $\theta$ before the experiment is run via a prior, $\prior$, which is usually assumed to be independent of our chosen design.
We then construct a simulator, $p(y|\theta,\xi)$, for possible experiment outcomes given $\theta$ and $\xi$.
Using this, we can concretely define the \emph{information gain} in $\theta$ from a hypothetical experiment as the reduction in Shannon entropy~\citep{shannon1948mathematical} from the prior to the posterior~\citep{lindley1956}:
\begin{align}
    \begin{split}
        \text{InfoGain}_{\theta}(\xi,y) &:= \mathrm{H}[\prior]-\mathrm{H}[\obspost] \\
        =\E_{\obspost}&\left[\log \obspost\right] - \E_{\prior}\left[\log \prior\right],
    \end{split}
    \label{eq:info_gain}
\end{align}
where $\obspost \propto \prior\obslik$.
As $y$ is unknown, we cannot optimize this directly and instead target the design that maximizes the \emph{expected information gain} (EIG)~\citep{lindley1956,bernardo1979expected,sebastiani2000maximum} in $\theta$, by using the marginal predictive distribution, $\obsmarg := \E_{\prior}[\obslik]$, to simulate outcomes:
\begin{align}
    \text{EIG}_{\theta}(\xi) := & ~\E_{\obsmarg}\left[\text{InfoGain}_{\theta}(\xi,y)\right] \nonumber           \\
    =                           & ~\E_{\prior\obslik}\left[ \log \obspost -\log\prior\right] \label{eq:eig-post} \\
    =                           & ~\E_{\prior\obslik}\left[ \log \obslik - \log\obsmarg\right].
    \label{eq:eig}
\end{align}
This EIG can be equivalently thought of as the mutual information between $y$ and $\theta$ given $\xi$, the expected reduction in predictive uncertainty from observing $\theta$, or the expected Kullback-Leibler (KL) divergence from the posterior to the prior.

Note that this framework does not require $\theta$ to correspond to some explicit model parameters (though this is still the most common scenario): we can define an EIG for almost any quantity of interest.
For example, we might be trying to learn about the optimum of a function~\citep{wang2017max,hennig2012entropy}, the output of an algorithm~\citep{neiswanger2021bayesian}, or future downstream predictions~\citep{bickfordsmith2023epig,kleinegesse2021gradientbased,mackay1992information}.
We may also have so-called nuisance parameters that are needed for modeling, but which we are not interested in gathering information on~\citep{chaloner1995}.

In all such cases, $p(\theta)$ and/or $p(y|\theta,\xi)$ will be defined \emph{implicitly}, rather than via a closed-form generative model with a known density function.
At the most basic level, the only thing that is needed to define an EIG is some process for jointly simulating $(\theta,y)$ pairs, from which everything else is uniquely defined.
A typical example scenario here is to have some underlying generative model, $p(\psi)p(y|\psi,\xi)$, for our experiment, with $\theta$ then defined either as a deterministic function of the parameters, $\psi$, or through an additional conditional sampling process, $p(\theta|\psi)$, that is independent of $y$.

Metrics of optimality other than the EIG are also possible~\citep{chaloner1995,ryan2016review,bernardo1979expected}, with the framework of BED more generally referring to any approach that optimizes an objective of the form $\E_{\prior\obslik}\left[U(\xi,\theta,y)\right]$ for some utility, $U$, that is a functional of the posterior, $\obspost$ (with some authors further relaxing this constraint on the form of $U$).
For example, the notion of an expected Fisher information gain has also recently been considered~\citep{prangle2022bayesian,overstall2022properties,walker2016bayesian}, due to its ability to be easily estimated without evaluating either the posterior or marginal likelihood.
Our focus, though, will be on maximizing the EIG defined in~\eqref{eq:eig-post}, as this is the most commonly used, and typically best-performing, approach.
Except when otherwise stated, we implicitly refer to this specific approach when using the term BED in the rest of the paper.

\subsection{Bayesian Adaptive Design}
\label{sec:BAD}

BED can be extended to adaptive settings---which we will refer to as \emph{Bayesian adaptive design} (BAD)---by iteratively incorporating information gathered during the experiment.
In particular, if our designs and outcomes can be broken down into a sequence of steps, denoted $\xi=\{\xi_1,\dots,\xi_T\}$ and $y=\{y_1,\dots,y_T\}$ respectively, then we can consider the \emph{incremental} EIG for each experiment step given the history, $h_{t-1}=\{(\xi_k,y_k)\}_{k=1}^{t-1}$:
\begin{align}
    \label{eq:cond-eig}
    \begin{split}
        &\text{EIG}_{\theta}(\xi_t | h_{t-1}) :=\\
        &\quad\E_{p(\theta|h_{t-1}) p(y_t|\theta, \xi_t, h_{t -1})} \left[ \log \frac{p(y_t|\theta, \xi_t, h_{t -1})}{p(y_t|\xi_t, h_{t-1})}\right],
    \end{split}
\end{align}
with $h_0=\varnothing$.
This can also be viewed as a conventional EIG with an updated prior and likelihood (in most cases $y_t$ is independent to $h_{t -1}$ given $\theta$ and $\xi_t$, such that only the prior needs to be updated).

The traditional BAD approach now chooses designs by optimizing $\text{EIG}_{\theta}(\xi_t | h_{t-1})$ at each experiment step~\citep{myung2013,watson2017quest+,chaloner1995,sebastiani2000maximum,ryan2014towards}, as per~\Cref{fig:boed_approaches_traditional}.
This provides a quantitative mechanism for adaptive experimentation.
It is usually more effective than \emph{static} design approaches---wherein all the $\xi_1,\dots,\xi_T$ are simultaneously chosen upfront---as information from previous observations is incorporated in each design decision through the iterative model updates.

\input{figures/bad.tex}

Note here that updates of the model between steps of BAD do not need to be restricted to updating our beliefs about $\theta$.
In particular, one should update beliefs for all the parameters of the underlying model (i.e.~$\psi$ in the earlier example) if these are distinct to $\theta$, with $p(\theta|h_{t-1})$ and $p(y_t|\theta, \xi_t, h_{t -1})$ implicitly defined through this update.
Failing to do this would lead to ignoring available information from previously gathered data in future design decisions.
Updating all the parameters also ensures that the sum of the incremental EIGs is equal to the total EIG across experiment iterations (see Equation~\eqref{eq:add_teig}).

\subsection{Why Take a Bayesian Approach?}

At this point, the reader might quite reasonably ask why one should take a Bayesian approach to experimental design instead of a classical frequentist one~\cite{ryan2007modern,fisher1937design,smith1918standard,atkinson2007optimum}.
At a high level, perhaps the biggest advantage is that, unlike comparable frequentist approaches, BED offers us a unified and self-consistent framework that incorporates all available information and does not rely on asymptotic approximations or model restrictions.
This advantage is particularly clear in adaptive settings, as here we require sequential decision-making in the presence of \emph{incomplete information}.
There is thus an inherent need to not only incorporate uncertainty into our decision making, but also to ensure these decisions are made in a way that is consistent with how our model will be updated at each iteration.
This is ensured by the self-similarity of the BAD framework, but is generally violated by frequentist approaches.

To explain the relative advantages of the BED approach more precisely, we first note that comparable classical frequentist approaches for optimizing designs are most commonly based on maximizing some functional of the Fisher information matrix (FIM) \cite{fisher1926arrangement,fisher1937design,smith1918standard,tsutakawa1972design,atkinson2007optimum},
\begin{align*}
    \text{FIM}(\theta,\xi) = \E_{p(y|\theta,\xi)} \left[
        \frac{\partial \log p(y|\theta,\xi)}{\partial \theta}
        \frac{\partial \log p(y|\theta,\xi)}{\partial \theta}^\top
        \right],
\end{align*}
where $p(y|\theta,\xi)$ is an assumed model that is equivalent to the likelihood in BED.
The use of the FIM here is motivated by its connections to the variance of estimators for $\theta$, but it also suffers from some significant drawbacks.

First, the FIM is a (square) matrix with a column/row for each dimension of the parameters.
As such, it cannot be directly used as an objective for optimizing designs.
Instead, one must optimize a summary statistic, such as its trace or determinant, leading to the classical `alphabetic criteria' for design optimality~\citep{atkinson1975design,elfving1952optimum,covey1970optimal,box1982}.
Unfortunately, these summary statistics do not necessarily fully reflect the gains in the joint distribution and, unlike the EIG, their behavior can vary with the precise parameterization of the model (though certain formulations are invariant to this~\citep{firth1997parameter}).

Second, and perhaps more importantly, the FIM is a function of the unknown ``true'' parameters, $\theta$.
This can severely undermine its practical deployment, given that we are gathering data to try and learn about $\theta$ in the first place.
Its use was originally motivated in the case of linear Gaussian models, for which it has the elegant property of being independent of $\theta$~\citep{fisher1937design}.
However, this is not the case in general, so users must typically either
(a) use a relaxation/approximation of the original model~\citep{hughes1998optimal};
(b) use some point estimate for the parameters (e.g.~a maximum likelihood estimate derived from existing data~\citep{atkinson2007optimum} or a worst-case configuration~\citep{chen2015minimax});
or (c) average over some prior on the parameters~\citep{chaloner1995,tsutakawa1972design,firth1997parameter,atkinson2007optimum,chaloner1989optimal}.

Here (a) will inevitably introduce errors into our design choices and will only be appropriate in the particular cases where this error can be kept small.

Meanwhile, (b) relies on point estimates for the very quantity we are trying to learn about.
This rather contradictory formulation means that it fails to properly account for our existing uncertainty and only uses the local properties of the model at the chosen parameter configuration.
This, in turn, can inevitably lead to sub-optimal design decisions that do not make use of all the information available.
It is particularly problematic in adaptive settings, as the FIM is only additive for multiple observations when the parameter estimate and likelihood model are held fixed, creating inconsistencies in the sequential decision-making process when iteratively incorporating data.

Though (c) can provide a practical approach, here one has now fully specified a Bayesian generative model, such that the approach is already arguably Bayesian anyway.
In fact, it has been demonstrated that some variations of this approach can actually be formulated under the general BED framework with a particular utility based on the gain in Fisher information~\citep{walker2016bayesian,prangle2022bayesian}.
More generally, Bayesian updates of the underlying distribution on $\theta$ are still required to allow consistent sequential decision making.

Putting these together, we thus see that, through their use of the FIM, classical frequentist approaches to experimental design either require us to make undesirable approximations, ignore our uncertainty in $\theta$ despite this being our target for learning, and/or leverage aspects from the BED framework itself.

Nonetheless, the EIG-based BED approach to experimental design is not without its shortfalls.
In particular, it can be extremely computationally challenging to conduct effectively, as we will discuss in the next section.
Further, its effectiveness is also inevitably tied into how well our underlying model matches the true data-generating process.
This model dependency is arguably unavoidable---as we are ultimately reasoning about data we have not yet gathered---but in practice it can be better to instead resort to heuristic approaches when constructing an accurate model is not possible.
In particular, even when using the EIG as a design objective in BAD, it can still sometimes be necessary to update the model in a non-Bayesian way when new data is acquired, as is common practice in the Bayesian active learning literature~\citep{settlestr09,gal2017deep}.

\section{A Computational Revolution}
\label{sec:estimation}

Though BED has principled information-theoretic foundations, it also poses some substantial computational challenges in the form of estimating and optimizing the EIG.
Historically, these have arguably formed the most significant barrier to its uptake, especially in the BAD context where decisions must often be made quickly.
We now outline these challenges and explain how our mechanisms for overcoming them have improved dramatically in recent years, transforming the applicability of BED.
We will assume for now that our model is explicit---i.e.~that we can evaluate the densities $p(\theta)$ and $p(y|\theta,\xi)$ in closed form---returning to the case where this is not true in Section~\ref{sec:implicit}.

\subsection{Nested Estimation}

Perhaps the most immediate computational challenge is that, except in special cases, the EIG is a doubly-intractable quantity that is difficult to estimate, let alone optimize.
In particular, it cannot be directly estimated using a conventional Monte Carlo estimator.

The origin of this issue is that, in general, both $\obsmarg$ and $\obspost$ are themselves intractable.
Thus, regardless of whether we use the form of the EIG from~\eqref{eq:eig-post} or~\eqref{eq:eig}, we have a term inside the expectation that is both intractable and varies between realizations of $y$.
For example, suppose we try to construct a Monte Carlo estimator of~\eqref{eq:eig}:
\begin{align}
    \label{eq:eig_mc}
    \text{EIG}_{\theta}(\xi) \approx \frac{1}{N}\sum_{n=1}^N \log p(y_n|\theta_n,\xi) - \log p(y_n|\xi),
\end{align}
where $\theta_n,y_n \iid p(\theta)p(y|\theta,\xi)$.
We now find that our estimator cannot itself be directly evaluated as the individual $p(y_n|\xi)$ terms are intractable.

If $y$ can only take on a small number of discrete values, $\mathcal{Y}$, this does not form too much of an issue as we can simply enumerate them, estimate $p(y|\xi)$ for each, and use a Rao-Blackwellized estimator~\citep{vincent2017darc,rainforth2018nesting,gal2017deep,foster2021variational}, such as
\begin{align}
    \label{eq:eig_finite}
    \begin{split}
        \hat{\mu}_N := \sum_{y\in\mathcal{Y}} &\,\frac{1}{N}\sum_{n=1}^N p(y|\theta_n,\xi) \log p(y|\theta_n,\xi) \\
        &- \hat{p}(y|\xi) \log \hat{p}(y|\xi),
    \end{split}
\end{align}
where $\hat{p}(y|\xi) = \frac{1}{N}\sum_{n=1}^N p(y|\theta_n,\xi)$.
However, if $y$ is continuous or the set $\mathcal{Y}$ is large, this is not feasible and we must fall back to sampling $y$, as per~\eqref{eq:eig_mc}.
Here estimation of the EIG for a given $\xi$ requires an additional \emph{nested} mechanism for estimating $p(y_n|\xi)$ for each $y_n$ \citep{hamada2001finding}, or, alternatively, approximating $p(\theta_n|y_n,\xi)$ for each pair $(\theta_n,y_n)$.

Traditional approaches for this nested estimation have typically either been based on nested Laplace approximations~\citep{lewi2009sequential,cavagnaro2010adaptive,long2013,ryan2015fully,long2022multimodal} or \emph{nested Monte Carlo} (NMC) estimators~\citep{rainforth2017thesis,rainforth2018nesting,hamada2001finding,ryan2003estimating,cook2008optimal,ryan2014towards,myung2013,beck2018fast,zheng2020sequential}.

While the former is prone to introducing substantial biases, the latter can provide asymptotically consistent estimators.
For example, the basic NMC estimator~\citep{rainforth2018nesting}
\begin{align}
    \label{eq:nmc}
    \munmc:= \frac{1}{N}\sum_{n=1}^N \log \frac{p(y_n|\theta_n,\xi)}{\frac{1}{M}\sum_{m=1}^M p(y_n|\theta_m',\xi)},
\end{align}
where $\theta_n,y_n \iid p(\theta)p(y|\theta,\xi)$ and $\theta_m' \iid p(\theta)$, has an asymptotic mean squared error to $\text{EIG}_{\theta}(\xi)$ of $\mathcal{O}(a/N+b/M^2)$ for constants $a$ and $b$ (see~\citep{rainforth2018nesting}), and thus converges as $N,M \to \infty$ (subject to some weak assumptions).

One can improve this estimator by replacing the simple Monte Carlo estimates for $p(y_n|\xi)$ with importance sampling ones, yielding the more general estimator~\citep{foster2019variational}
\begin{align}
    \label{eq:nmcq}
    \munmcq:= \frac{1}{N}\sum_{n=1}^N \log \frac{p(y_n|\theta_n,\xi)}{\frac{1}{M}\sum_{m=1}^M \frac{p(y_n|\theta_m',\xi)p(\theta_m')}{q(\theta_m'|y_n,\xi)}},
\end{align}
where we now draw the inner samples from our proposal: $\theta_m' \iid q(\theta_m'|y_n,\xi)$.
This can help reduce the constants in the convergence rate (i.e.~$a$ and $b$) compared with $\munmc$, along with the finite sample bias and variance.

Unfortunately, though, all NMC estimators have a number of undesirable properties.
First, they are, in general, biased for any finite value of $M$, due to the fact that nonlinear mappings of unbiased estimators are themselves biased estimators.
Second, and perhaps more importantly, they can be very expensive to evaluate as their cost scales as $C=NM$.
In turn, this causes them to have a slow convergence rate: for most problems, they cannot achieve a better mean squared error rate than $\mathcal{O}(C^{-2/3})$, with this occurring when $M\propto \sqrt{N}$~\citep{rainforth2018nesting}.

\subsection{Debiasing Schemes}

An important recent advancement in BED has been the development of mechanisms for reducing~\citep{goda2020multilevel}, or even eliminating~\citep{goda2022unbiased}, the bias of EIG estimators and their gradients.
While debiasing techniques for (nested) Monte Carlo estimators date back to at least~\citep{mcleish2011general,rhee2015unbiased,lyne2015russian} (with some aspects even having much earlier origins in the physics literature~\citep{kennedy1985noise,bhanot1985bosonic}), their application to the specific problem of EIG estimation is a more recent development.

The basis for these debiased EIG estimators is the framework of multi-level Monte Carlo~\citep{giles2008multilevel}.
Perhaps most notably, Goda et al. (2022)~\citep{goda2022unbiased} recently showed that the multi-level Monte Carlo variant of~\citep{rhee2015unbiased} can be used to generate fully \emph{unbiased} estimates of the EIG and its gradients.
Their approach works by first expressing the EIG as the expectation of the NMC estimator in~\eqref{eq:nmcq} with $N=1$ and $M=\infty$, then noting that this can be written as a telescopic sum.
Namely,
\begin{align}
    \text{EIG}_{\theta}(\xi) & = \E[\munmcq[1][\infty]] =  \E\left[\sum\nolimits_{\ell=0}^\infty \Delta_\ell \right],
\end{align}
where $\Delta_0 := \munmcq[1][M_0]$ for a user-chosen integer $M_0$, and
\begin{align}
    \Delta_\ell := \munmcq[1][M_0 2^{\ell}]-\frac{1}{2}\left(\munmcq[1][M_0 2^{\ell-1}]^{(a)}+\munmcq[1][M_0 2^{\ell-1}]^{(b)}\right)
\end{align}
for $\ell>0$, where the samples of $\theta'$ used in $\munmcq[1][M_0 2^{\ell}]$ are split equally between $\munmcq[1][M_0 2^{\ell-1}]^{(a)}$ and $\munmcq[1][M_0 2^{\ell-1}]^{(b)}$, forming an antithetic coupling~\citep{giles2014antithetic} through the sharing.
They then introduce an importance sampler to produce an unbiased estimate of the infinite sum $\sum_{\ell=0}^\infty \Delta_\ell$ with a single sampled term, such that we have
\begin{align}
    \label{eq:mlmc_est}
    \text{EIG}_{\theta}(\xi)
     & = \E\left[\Delta_{\ell} / r(\ell) \right],
\end{align}
where the expectation is now over both $\ell \sim r(\ell)$ and $\Delta_{\ell}$.
The EIG can now be unbiasedly estimated through direct Monte Carlo estimation of~\eqref{eq:mlmc_est}.

Critically, the antithetic coupling used means that it is possible, under weak assumptions, to construct $r(\ell)$ such that the resulting estimators (and their gradients with respect to $\xi$) have both finite expected variance and cost.
Specifically,~\citep{goda2022unbiased} show that these properties hold for most problems when we take $r(\ell)\propto 2^{-\tau \ell}$ for any $1<\tau<2$.
While the cost of the resulting estimators can still be significant (as each sample of $\Delta_{\ell}$ itself requires $M_0 2^{\ell}+1$ likelihood evaluations), it can provide noticeable theoretical and practical gains.
In particular, it allows convergence rates $\mathcal{O}(C^{-1/2})$ in the total cost, $C$, to be achieved, thereby recovering the standard (unnested) Monte Carlo rate.

\subsection{Functional and Variational Approaches}

Another important area of recent progress is in the development of approaches that circumvent the inefficiency of NMC estimators by utilizing some kind of functional approximation for the intractable nested term~\citep{kleinegesse2018efficient,foster2019variational,foster2020unified,foster2021dad,ivanova2021implicit,huan2013simulation,overstall2020bayesian,kennamer2022design}.
Unlike the aforementioned classical Laplace-based approximations, many of these new techniques---which are often based around variational approximations~\citep{foster2019variational}---can provide consistent EIG estimates, accommodate implicit likelihood models, and improve synergy with the design optimization process itself.

To explain the motivations for these approaches, consider the form of the EIG in~\eqref{eq:eig}, where the nested term we need to estimate is $p(y|\xi)$.
Rather than estimating it from scratch each time we sample a new $y_n$, we can note that it will usually have a degree of smoothness that we can exploit to transfer information between estimates.
In fact, we can view the problem as a regression from $y$ to $p(y|\xi)$, where our estimates of $p(y|\xi)$ provide unbiased noisy evaluations of the target.
Alternatively, we can also think about estimating $p(y|\xi)$ using a simple density estimator trained on samples of $y$ (utilizing the fact we can draw from the joint, $p(\theta)p(y|\theta,\xi)$).
Either way, our learned approximation, $q(y|\xi)$, can then be used to directly replace the intractable $p(y|\xi)$ term in~\eqref{eq:eig_mc}, after which we can estimate the EIG via standard Monte Carlo.

This two-stage process can provide substantial computational savings: the costs from approximating $p(y|\xi)$ and the subsequent Monte Carlo estimation are now additive, not multiplicative.
Indeed,~\citep{foster2019variational} shows that such an approach can theoretically recover convergence rates of $\mathcal{O}(C^{-1/2})$ in total cost $C$ (though, in practice, there are usually limits on how exact a functional approximation we can learn).

\subsubsection{Variational Approaches}
\label{sec:variational}

An intriguing property of this framework is that it turns out that approximating $p(y|\xi)$ with $q(y|\xi)$ actually produces a \emph{variational bound} on the EIG whenever the latter forms a valid normalized density~\citep{foster2019variational}, such that
\begin{align}
    \text{EIG}_{\theta}(\xi)
     & \le \E_{\prior\obslik}\left[ \log \obslik - \log q(y|\xi)\right],
    \label{eq:var_marg}
\end{align}
with equality holding when $\forall y: q(y|\xi)=p(y|\xi)$.
This variational bound is distinct from those on the log evidence commonly used by the approximate inference~\citep{blei2016variational,hoffman2013stochastic} and variational auto-encoder communities~\citep{kingma2014adam,rezende2014stochastic,burda2015importance}, but it behaves in very much the same way and can be used as an objective for optimizing our approximation $q(y|\xi)$.

In fact, one can construct a whole host of different variational bounds for the EIG by approximating different terms in the expectation.
For example, if we introduce an amortized inference network~\citep{stuhlmuller2013learning,kingma2014auto}, $q(\theta|y,\xi)$, that maps from outcomes to approximations of the corresponding posterior, $p(\theta|y,\xi)$, then we instead get the variational \emph{lower} bound~\citep{foster2019variational,pacheco2019variational,barber2003imalgorithm}
\begin{align}
    \text{EIG}_{\theta}(\xi)
     & \ge \E_{\prior\obslik}\left[ \log q(\theta|y,\xi) - \log p(\theta)\right],
    \label{eq:var_post}
\end{align}
with equality again holding when the approximation is exact.
Even the expectation of the NMC estimator in~\eqref{eq:nmcq} turns out to itself be a variational upper bound~\citep{foster2019variational},
\begin{align}
    \label{eq:vnmc}
    \text{EIG}_{\theta}(\xi) & \le \E_{\prior\obslik}\left[\munmcq[1][M]\right],
\end{align}
which can be arbitrarily tightened by increasing $M$.
In both these cases we can also use the learned $q(\theta|y,\xi)$ as a proposal for an efficient NMC estimator using~\eqref{eq:nmcq}, allowing for further reduction of the estimator bias if needed.

Note that these are not the only variational bounds one can construct (see, e.g.,~\citep{foster2019variational,foster2020unified,ivanova2021implicit,foster2021variational,kleinegesse2021gradientbased} for additional examples).
Furthermore, as the EIG is equivalent to the mutual information between $y$ and $\theta$, a number of methods for optimizing mutual information outside of the experimental design setting can also be applied~\citep{poole2018variational,belghazi2018mutual,oord2018representation,guo2021tight}.
Indeed,~\eqref{eq:var_post} is exactly equivalent to the classical mutual information bound of~\citep{barber2003imalgorithm}.

\subsubsection{Estimation for Implicit Models}
\label{sec:implicit}

Another important class of functional-approximation approaches for EIG estimation are based on accommodating implicit likelihood models, wherein we can draw samples for $y|\theta,\xi$, but the density $p(y|\theta,\xi)$ cannot be evaluated in closed form.
This introduces an additional intractable term that must be approximated and prevents many of the previously introduced estimators from being used directly.

Some approaches have been introduced that approximate this density in isolation~\citep{huan2013simulation,overstall2020bayesian}, while~\citep{kleinegesse2018efficient} use the approach of~\citep{thomas2022likelihood} to estimate the ratio $p(y|\theta,\xi)/p(y|\xi)$ using logistic regression.
Variational bounds that allow for implicit likelihood models (e.g.~\eqref{eq:var_post}) have been also developed~\citep{kleinegesse2020mine,ivanova2021implicit,foster2019variational,foster2020unified}, inspired by analogous bounds used in the representation learning community~\citep{belghazi2018mutual,oord2018representation,nguyen2010nwj}.
Historically, approximate Bayesian computation techniques have also been used to deal with implicit likelihoods in a BED context~\citep{drovandi2013bayesian,Hainy2016likelihoodfree,Price2016,dehideniya2018}.

Dealing with implicit priors is typically easier than implicit likelihoods, as one can use formulations based on~\eqref{eq:eig} that avoid calculating the prior density (e.g.~\eqref{eq:nmc}, \eqref{eq:var_marg}).
While some of the recently developed variational bounds reintroduce a need to calculate the prior density (e.g.~\eqref{eq:var_post}, \eqref{eq:vnmc}), there are some that do not, critically including some that also allow implicit likelihood models as well~\citep{kleinegesse2020mine,ivanova2021implicit}.

\subsection{Optimization}

The ultimate aim in BED is to produce the best possible designs: the EIG is an objective that we wish to optimize, rather than simply estimate.
Historically, many of the most commonly used optimization schemes for this have acted as wrappers around the estimator.
Examples of this include simple discretization and enumeration schemes~\citep{carlin1998approaches,ryan2003estimating,watson2017quest+,van2003optimal,lyu2019ultra,han2004bayesian,gillespie2019efficient} (sometimes coupled with dynamic resource allocation~\citep{rainforth2017thesis,vincent2017darc}), Bayesian optimization~\citep{foster2019variational,kleinegesse2018efficient,von2019optimal}, evolutionary algorithms~\citep{hamada2001finding,price2018induced}, and co-ordinate exchange methods~\citep{meyer1995coordinate,overstall2017bayesian}.

Unfortunately, all these approaches can be extremely slow to run, as they require many individual EIG estimations to be performed during the optimization.
They also often scale poorly to high-dimensional design spaces, as they have limited ability to exploit problem-specific information like gradients.

Others have suggested sampling-based approaches that couple the outer sampling from $p(\theta)p(y|\theta,\xi)$ with the optimization of $\xi$~\citep{ryan2014towards,clyde1996exploring,drovandi2013bayesian}, often employing extended spaces to concentrate sampling around the optimum of the EIG~\citep{amzal2006bayesian,muller2004optimal,muller2005simulation,cook2008optimal,kuck2006smc,stroud2001optimal}.
These approaches can sometimes provide efficiency and scaling benefits over more black-box optimizers, but are still typically quite expensive procedures.
They further tend to require customized proposals to be constructed, which can be challenging, especially for high-dimensional problems.

\subsubsection{Stochastic Gradient Schemes}
\label{sec:sga}

In principle, optimizers based on stochastic gradients~\citep{robbins1951stochastic} provide the potential for huge improvements in efficiency and scaling over such approaches by side-stepping approximation of the EIG and instead directly leveraging small-sample estimates of its derivative with respect to $\xi$.
Such a stochastic gradient approach was first suggested for EIG optimization by Huan and Marzouk (2014)~\citep{huan2014gradient} using gradients of the basic NMC estimator given in~\eqref{eq:nmc}.
Their method was more recently revisited by Carlon et al.~(2020)~\citep{carlon2020nesterov}, who also consider using stochastic gradient ascent with nested Laplace approximations and the more general NMC estimator given in~\eqref{eq:nmcq}.

Unfortunately, these approaches introduce bias into the optimization due to the bias in their underlying estimators, potentially leading to sub-optimal designs in turn.
This, though, is where the full importance of the aforementioned developments in debiasing schemes and variational bounds really becomes apparent: both can provide a consistent method for optimizing the EIG using stochastic gradients.

Foster et al.~(2020)~\cite{foster2020unified} introduced such a unified stochastic gradient approach to BED by simultaneously maximizing variational \emph{lower} bounds to the EIG (e.g.~\eqref{eq:var_post} but not~\eqref{eq:var_marg} or~\eqref{eq:vnmc}) with respect to both the variational and design parameters.
They also show that the small adjustment to~\eqref{eq:vnmc} of including the outer $\theta$ sample in the estimate of $p(y|\xi)$ changes it from an upper to a lower bound, so that it can be used for stochastic gradient ascent:
\begin{align}
    \label{eq:ace}
    \text{EIG}_{\theta}(\xi) \ge \E\left[\log \frac{p(y|\theta_0,\xi)}{\frac{1}{M+1}\sum_{m=0}^M \frac{p(y|\theta_m,\xi)p(\theta_m)}{q(\theta_m|y,\xi)}}\right],
\end{align}
where $\theta_0,y \sim p(\theta)p(y|\theta,\xi)$ and $\theta_m | y \iid q(\theta_m|y,\xi)$ for $m\ge 1$.
This contrastive bound again recovers the true EIG when $q(\theta|y,\xi)$ matches the true posterior, while allowing the tightness of the bound to be controlled through $M$.
It is also useful for evaluation, as it can be used in conjunction with~\eqref{eq:nmcq} to provide upper and lower limits on the true EIG.

Foster et al.~(2020)~\cite{foster2020unified} further showed that~\eqref{eq:ace} remains a valid lower bound when the likelihood density, $p(y|\theta,\xi)$, is replaced by any (unnormalised) approximation thereof.
This provides a mechanism for working with implicit likelihood models (as an alternative to~\eqref{eq:var_post}, which can still be used directly) by simultaneously training this approximation alongside the variational posterior approximation and design itself.
In independent work published shortly after,~Kleinegesse and Gutmann (2020)~\cite{kleinegesse2020mine} showed that the mutual information bound of~\citep{belghazi2018mutual} can also be used in an analogous manner, allowing the posterior and likelihood approximations to be collapsed to a single `critic' network.
Unlike~\citep{foster2020unified}, they go on to provide experimental demonstrations that their approach can practically be used in implicit likelihood settings.
Further bounds and unified stochastic gradient approaches have since been developed for both explicit and implicit models by~\citep{kleinegesse2021gradientbased,zhang2021sagabed,foster2021dad,ivanova2021implicit,guo2021tight,prangle2022bayesian}.

The aforementioned multi-level Monte Carlo approach of~\citep{goda2022unbiased} alternatively provides unbiased, finite-variance estimates of the EIG gradient, $\nabla_{\xi} \text{EIG}_{\theta}(\xi)$, which can thus be directly used to perform stochastic gradient ascent.
This avoids the need for any variational distribution---and thus any limitations originating from the variational family---but results in a higher per-sample cost and may increase estimator variance; it has not yet been empirically compared to variational approaches in the literature.

An inevitable problem with all these gradient-based approaches occurs when the design space or likelihood function is not continuous, for example because some design decisions are discrete.
Relaxation schemes can, in principle, allow such cases to be dealt with~\citep{ivanova2022efficient,zhang2021sagabed}, but further work is still required to fully address this problem.

\section{From Designs to Policies}
\label{sec:policy}

Despite all these innovations, BAD (i.e.~sequential BED) can still be a prohibitively expensive endeavor.
In particular, even if our EIG optimization is extremely efficient, the traditional BAD approach discussed in Section~\ref{sec:BAD} still needs to update the model at each step via Bayesian inference, which is typically itself expensive.
The approach is thus only practical when such computation can be justified \emph{during} the experiment, which is often not the case.

Further, the traditional BAD approach is actually subtly sub-optimal because it makes greedy, myopic, decisions: it fails to account for the fact that different design choices at one step can also influence the information gained from future steps~\citep{borth1975total,golovin2010near,ryan2016review,foster2021variational}.
Approaches to try and overcome this have been proposed in the Bayesian optimization literature~\citep{gonzalez2016glasses,jiang2020binoculars}, but they add significant further cost.

Remarkably, recent developments have shown that it is actually possible to simultaneously overcome both of these shortfalls by reimagining the traditional BAD framework itself~\citep{huan2016sequential,foster2021dad,ivanova2021implicit,foster2021variational,blau2022optimizing}.
Rather than optimizing the designs directly, these methods instead use upfront training to learn a \emph{policy} that can then be deployed to make design decisions for us based on the previously gathered data.

This design policy concept was first proposed by Huan and Marzouk (2016)~\citep{huan2016sequential} using ideas from dynamic programming and reinforcement learning.
The policies they learn map from explicit representations of the posterior (which form the RL `state') to the next design decision.
As such, significant computation is still required at each iteration in the form of the posterior updates themselves.

\subsection{Deep Adaptive Design}

An exciting new policy-based framework---called \emph{deep adaptive design} (DAD)---was recently proposed in~\citep{foster2021dad,ivanova2021implicit} that instead avoids the need for \emph{any} significant computation to be performed during the experiment, while also allowing superior, non-myopic, automated policies to be learned.

The high-level idea of DAD is summarized in~\Cref{fig:boed_approaches_idad}.
While the traditional BAD approach (cf~\Cref{fig:boed_approaches_traditional}) requires costly inference and optimization at each step of the experiment, DAD uses the model to learn a \emph{policy network}, $\designnet$, upfront.
Unlike the policies in~\citep{huan2016sequential}, $\designnet$ maps directly from the previously collected data to the next design decision, side-stepping the need to perform inference during the experiment.
Specifically, it is a deep neural network with parameters $\phi$ that takes in a history of varying length, $h_{t-1}$, and returns the next design to evaluate: $\xi_t=\designnet(h_{t-1})$.
It thus allows design decisions to be made using a single forward evaluation of the network.

This forward evaluation is generally very cheap, allowing design decisions to be made almost instantly and thus adaptation in \emph{real time}.
Further, once $\designnet$ is learned, it can be directly applied to multiple realizations of the experiment (e.g. different participants in a survey), allowing for the cost of adaptation to be amortized.

On top of these transformative speed gains, DAD has also been empirically found to provide significant improvements in the quality of designed decisions made compared to traditional greedy strategies~\citep{foster2021dad,ivanova2021implicit,blau2022optimizing}.
This has been attributed to a combination of it learning non-myopic policies and avoiding the errors caused by the approximate inference schemes required by previous frameworks.

\input{figures/dad.tex}

\subsection{Learning Policies}

The underlying concept of a design policy from these works is extremely general and a strict generalization of the traditional BAD approach, which can be viewed trying to approximate the implicit policy $\pi_{\text{trad}}(h_{t-1}):= \text{argmax}_{\xi_t} \text{EIG}_{\theta}(\xi_t|h_{t-1})$.
Note here that we can always express our policy as a function of the history without loss of generality: for a given underlying model, all available information relevant to making a design decision is contained in the history.
For example, the posterior belief states used by both $\pi_{\text{trad}}$ and the policies of~\citep{huan2016sequential} are entirely determined by the model and history.
Thus DAD's choice of learning policies that directly map from histories to designs has clear justification.

As previously mentioned, $\pi_{\text{trad}}$ is generally a sub-optimal policy because of its greedy decision making~\citep{borth1975total}.
When learning a policy, we thus instead want to approximate the true optimal policy, which is the policy that maximizes the \emph{total} EIG over all $T$ experiment steps~\citep{huan2016sequential,foster2021dad,ivanova2021implicit}:
\begin{align}
    \hspace{-1pt}\text{TEIG}_{\theta}(\designnet) & \!:=\!
    \E_{\prior p(y_{1:T} | \theta, \designnet)} \left[\log \frac{p(y_{1:T} | \theta, \designnet)}{p(y_{1:T} | \designnet)} \right],
    \label{eq:total_eig}
\end{align}
where $p(y_{1:T} | \designnet) = \E_{\prior} \left[p(y_{1:T} | \theta, \designnet)\right]$ and $p(y_{1:T}|\theta,\designnet)$ is autoregressively defined by iteratively evaluating $\xi_t = \designnet(h_{t-1})$, sampling $y_t\sim p(y_t|\theta,\xi_t,h_{t-1})$, and assigning $h_t = h_{t-1} \cup (\xi_t,y_t)$.

Note here that the individual incremental EIGs are additive in expectation, so the total EIG is also equal to the sum of the EIGs from each experiment iteration~\citep{foster2021dad,ivanova2021implicit}:
\begin{align}
    \label{eq:add_teig}
    \begin{split}
        \text{TEIG}_{\theta}&(\designnet) = \text{EIG}_{\theta}(\designnet(\varnothing))\\
        &+\sum_{t=2}^{T} \E_{p(y_{1:t-1}|\designnet)} \left[\text{EIG}_{\theta}(\designnet(h_{t-1})|h_{t-1})\right].
    \end{split}
\end{align}

For a given model, the total EIG is only a function of the policy parameters, $\phi$, with the designs themselves becoming random variables (because of their dependency on previous outcomes) that are marginalized out in the objective.
We can therefore learn a policy by simply optimizing $\phi$ to maximize the total EIG, analogously to optimizing the designs in a static BED setting.

The original DAD paper~\citep{foster2021dad} showed how this optimization can be performed effectively using variational bounds coupled with stochastic gradient schemes---like those introduced in~\Cref{sec:variational} and~\Cref{sec:sga} respectively.
It further introduced a customized architecture for the policy network.
The iDAD approach of~\citep{ivanova2021implicit} later generalized this training scheme to implicit likelihood models and provided further refinements on the policy network architecture itself.
Others have also since built on the framework by instead using reinforcement learning approaches to learn the policy~\citep{blau2022optimizing,lim2022policy,foster2021variational}.

\section{Future Directions}

\subsection{Policy-Based BAD}

The recent emergence of policy-based BAD methods has led to a significant improvement in state-of-the-art performance in terms of both deployment speed and design quality~\citep{foster2021dad,ivanova2021implicit,blau2022optimizing}.
However, they still represent a fledgling approach and we believe their potential stretches far beyond what has been achieved so far.
We thus anticipate substantial opportunities in developing new policy-based approaches to improve and generalize current methods.

Perhaps the biggest challenge here, and indeed a key challenge for BED more generally, will be scaling to larger and more complex problems.
Though significant progress on scalability has already been made in recent years, current methods can still be limited by a number of factors, such as the dimensionality and smoothness of both the model and the design space, and the length of the experiment sequence, $T$.
Dealing with discrete design components can also be difficult.
We believe that it should be possible to make substantial improvements in our ability to deal with such challenges by making advancements on a number of different fronts, such as network architectures, objectives and estimators, and training mechanisms.

\subsection{Linking with Related Areas}

The BED literature is spread across a wide variety of different fields, with papers appearing in, for example, statistics~\citep{prangle2022bayesian,overstall2022properties}, machine learning~\citep{foster2021dad,blau2022optimizing}, engineering~\citep{carlon2020nesterov,beck2018fast}, physics~\citep{huan2013simulation}, and social science~\citep{myung2013,kasy2021adaptive} venues.
This somewhat scattered nature has sometimes led to a (potentially problematic) degree of detachment between research threads.
Moreover, some important links between BED and other research areas have recently started to become apparent, with these connections ready to be exploited.

Perhaps the most obvious related literature is that of Bayesian active learning~\citep{houlsby2011bayesian,gal2017deep,golovin2010near,mackay1992information,kossen_active_2020,kapoor2007active,kossen2022active}.
Bayesian active learning can be viewed as an important special case of BAD, wherein our aim is to adaptively train a machine learning model in a data-efficient manner by iteratively acquiring the required labels.
Here the designs correspond to the inputs we acquire labels for, which are typically chosen from a finite pool of unlabeled datapoints.
This is most commonly done by following the traditional BAD approach outlined in Section~\ref{sec:BAD} (often with datapoints chosen in batches~\citep{kirsch2019batchbald,pinsler19bayesian}), with the EIG for the model parameters typically referred to as the \emph{Bayesian active learning by disagreement} (BALD) score in the active learning literature~\citep{houlsby2011bayesian}.

We believe that there is much to be gained in reestablishing links and transferring ideas between the intrinsically connected Bayesian active learning and general BED literatures.
For example, the Bayesian active learning community has had notable success in dealing with much larger and higher-dimensional datasets---like those seen in computer vision~\citep{gal2017deep,beluch2018power} and natural language processing~\citep{siddhant2018deep,osband2023finetuning,shen2018deep}---than have typically been considered elsewhere.
On the other hand, Bayesian active learning has seen limited application outside of classification problems with a modest number of classes, relying on enumeration of $y$ and restricted estimators like~\eqref{eq:eig_finite} to keep costs manageable.
Even for such cases, it can still experience computational difficulties when performing the large batch acquisitions often used in practice~\citep{kirsch2019batchbald,pinsler19bayesian}.

The recent computational developments discussed earlier are well-placed to address these shortfalls, while there is also obvious potential in applying policy-BAD approaches in the active learning setting.
More fundamentally, it has also recently been suggested that the BALD score is a sub-optimal application of BED for active learning~\citep{bickfordsmith2023epig}, as we care about maximizing information gain for downstream \emph{predictions}, not the model parameters.

Another under-explored area of overlap is between BAD and model-based approaches to reinforcement learning (RL)~\citep{sutton2018reinforcement,szepesvari2010algorithms}, especially Bayesian RL methods~\citep{igl2018deep,ghavamzadeh2015bayesian}.
Though few links have historically been drawn between the fields, the concept of a design policy has a clear analogy to the policies used in RL.
Indeed,~\citep{huan2016sequential} explicitly first introduced the idea of a BAD policy through an RL formulation, while others have also recently directly used RL techniques to learn design policies~\citep{blau2022optimizing,lim2022policy,shen2021bayesian}.

More specifically, BAD can actually be formulated as a Bayes adaptive Markov decision process~\citep{guez2012efficient,ross2007bayes,duff2002optimal} with a reward function based on the (incremental) EIG~\citep{foster2021variational,huan2016sequential}.
Thus, while there are clear differences in how policies are trained and utilized between BAD and Bayesian RL, there is also concrete overlap between the problems as well.
The concept of exploration in RL~\citep{foster2021variational,jorke2022simple} is also itself underpinned by the idea of information gain, with information-based reward signals analogous to the EIG previously used in RL contexts by~\citep{sun2011planning,schmidhuber2010formal}.
With these links now established, but largely unexplored, we believe that further cross-dissemination of ideas between the fields will be of substantial mutual benefit.

\subsection{Model Misspecification and Downstream Analysis}

Despite its theoretical elegance and asymptotic guarantees~\citep{paninski2005asymptotic}, BED, and in particular BAD, can suffer from serious pathologies in practice if our model is misspecified~\citep{kennedy2001bayesian,brynjarsdottir2014learning,grunwald2017inconsistency,kleijn2012bernstein,feng2015optimal}, that is there is no $\theta^*$ such that $p(y|\theta=\theta^*,\xi)$ matches the true distribution for $y|\xi$.

While such misspecification is an inevitable issue in any Bayesian setting, BED approaches can be particularly sensitive because we are not only using our model to fit the data, but also to gather new data as well.
Most problematically, because Bayesian approaches only capture uncertainty within the model, and not any uncertainty about whether the model itself is correct, one can occasionally see catastrophic failure cases where BAD becomes stuck querying similar designs, producing poor-quality datasets~\citep{vincent2017darc}.
An extreme example of this is given by linear regression models, where the EIG of the regression coefficients is always maximized at the extrema of the inputs, regardless of the chosen prior.
This is not necessarily an issue if the true relationship really is linear, but the subsequent failure to explore any of the interior of the design space is likely to be highly problematic if it is not.

While misspecification is not something we can realistically hope to eliminate completely, at least not in a general BED context, there is still much to be done to better understand and manage it.
At present, there is only limited literature on both its theoretical~\citep{farquhar2020statistical,fudenberg2017active,overstall2021bayesian} and empirical implications~\citep{sloman2022characterizing,feng2015optimal}, with little done to investigate the precise failure mechanisms that can occur.
Similarly, only a small number of practical methods for mitigating the effects of such misspecification have been suggested~\citep{overstall2021bayesian,go2022robust,ouyang2016,etzioni1993optimal}.
Given the potential dangers of model misspecification in BED, it is clear that more work is needed on both of these fronts.

An important related issue can occur when the gathered data is utilized for purposes other than updating the model used for data gathering.
Theoretically, the likelihood principle~\citep{berger1988likelihood,barnard1962likelihood} provides protection against downstream Bayesian analysis of the gathered data yielding erroneous conclusions, provided the model used for this analysis is not itself misspecified (even if the model used for data gathering is).
However, in practice, model misspecification in downstream analysis is just as unavoidable as during data gathering, and the data we gather can influence the impact of this misspecification.

Moreover, the data-gathering process can also significantly influence not only the informativeness of the data for downstream problems, but also the model selection and checking processes.
Gathered data is also sometimes used downstream for entirely non-Bayesian purposes, such as training a machine learning model via empirical risk minimization.
This is a scenario that regularly occurs in a Bayesian active learning context~\citep{farquhar2020statistical}, wherein accurate inference is often infeasible and one resorts to low-fidelity posterior approximations~\citep{pinsler19bayesian,kirsch2019batchbald,siddhant2018deep}, or even entirely non-Bayesian updates~\citep{beluch2018power,gal2017deep,margatina2021bayesian}.
More work is required to understand the influence of using BED to gather data in such contexts, and to develop approaches that ensure the data collected has good properties beyond just inference in the chosen model.

\subsection{Models and Applications}

As a model-based approach, the performance of BED is only ever as good as the underlying model.
Improvements in models themselves will therefore always be one of, if not the, most important vectors for advancing BED.
Though most such advancements will inevitably be problem-specific, there are still two important general points worth noting.

First, recent developments in BED for implicit likelihoods models~\citep{kleinegesse2020mine,ivanova2021implicit,overstall2020bayesian,foster2019variational} have significant potential implications for model development.
It is often much easier to construct an accurate \emph{simulator} for experiment outcomes than a classical closed-form likelihood.
Moreover, many such simulators already exist across the sciences, to model things from quantum logic gates~\citep{wittler2021integrated} to large-scale climate simulations~\citep{fanourgakis2019evaluation}.
Such simulators can be interpreted as implicit likelihood models~\citep{rainforth2017thesis}, and therefore modern approaches allow them to be used for BED.

Second, the aforementioned substantial recent computational advancements are potentially transformative in the complexity of the models that we can consider using with BED.
As such, practitioners should look to embrace this freedom and develop models that are flexible and accurate to their beliefs, rather than restricting themselves to simple models which allow easy estimation, as has often been the case in the past.
Further, we should now consider BED as a viable approach for larger and more complex problems than those it has most commonly previously been utilized for.
Indeed, finding the limits of these new approaches is an important area for future work in its own right.

\begin{acks}[Acknowledgments]
We would like to thank Dennis Prangle and Christian Robert for inviting us to write this paper and their helpful feedback during the review process.
We would also like to thank the additional anonymous reviewer for their useful suggestions.
\end{acks}

\begin{funding}
Desi R Ivanova was supported by the EPSRC Centre for Doctoral Training in Modern Statistics and Statistical Machine Learning (EP/S023151/1).

Freddie Bickford Smith was supported by the EPSRC Centre for Doctoral Training in Autonomous Intelligent Machines and Systems (EP/S024050/1).
\end{funding}

\bibliographystyle{imsart-number}
\bibliography{main}

\end{document}

%% file: figures/bad.tex
\begin{figure}[t]
	\centering
	\tikzset{every picture/.style={line width=0.75pt}}
	\begin{tikzpicture}[x=0.75pt,y=0.75pt,yscale=-0.85,xscale=1.1]
		\draw  [color={rgb, 255:red, 255; green, 255; blue, 255 }  ,draw opacity=1 ][fill={rgb, 255:red, 245; green, 166; blue, 35 }  ,fill opacity=0.53 ] (171.17,119.93) .. controls (171.17,116.01) and (174.35,112.83) .. (178.27,112.83) -- (229.07,112.83) .. controls (232.99,112.83) and (236.17,116.01) .. (236.17,119.93) -- (236.17,141.23) .. controls (236.17,145.15) and (232.99,148.33) .. (229.07,148.33) -- (178.27,148.33) .. controls (174.35,148.33) and (171.17,145.15) .. (171.17,141.23) -- cycle ;
		\draw  [color={rgb, 255:red, 255; green, 255; blue, 255 }  ,draw opacity=1 ][fill={rgb, 255:red, 208; green, 2; blue, 27 }  ,fill opacity=0.38 ] (231.5,196.03) .. controls (231.5,192.11) and (234.68,188.93) .. (238.6,188.93) -- (289.4,188.93) .. controls (293.32,188.93) and (296.5,192.11) .. (296.5,196.03) -- (296.5,217.33) .. controls (296.5,221.25) and (293.32,224.43) .. (289.4,224.43) -- (238.6,224.43) .. controls (234.68,224.43) and (231.5,221.25) .. (231.5,217.33) -- cycle ;
		\draw  [color={rgb, 255:red, 255; green, 255; blue, 255 }  ,draw opacity=1 ][fill={rgb, 255:red, 74; green, 144; blue, 226 }  ,fill opacity=0.35 ] (108.5,196.46) .. controls (108.5,192.54) and (111.68,189.36) .. (115.6,189.36) -- (166.4,189.36) .. controls (170.32,189.36) and (173.5,192.54) .. (173.5,196.46) -- (173.5,217.76) .. controls (173.5,221.68) and (170.32,224.86) .. (166.4,224.86) -- (115.6,224.86) .. controls (111.68,224.86) and (108.5,221.68) .. (108.5,217.76) -- cycle ;
		\draw   (229.72,208.39) -- (183.28,208.39) -- (183.28,211.73) -- (174.72,205.58) -- (183.28,199.43) -- (183.28,202.77) -- (229.72,202.77) -- cycle ;
		\draw   (141.31,185.78) -- (141.31,134.73) -- (160.41,134.73) -- (160.41,138.57) -- (169.2,132.17) -- (160.41,125.78) -- (160.41,129.62) -- (136.2,129.62) -- (136.2,185.78) -- cycle ;
		\draw   (238.58,133.28) -- (262.17,133.28) -- (262.17,180.16) -- (257.88,180.16) -- (264.73,188.16) -- (271.58,180.16) -- (267.29,180.16) -- (267.29,128.16) -- (238.58,128.16) -- cycle ;
		\draw  [color={rgb, 255:red, 255; green, 255; blue, 255 }  ,draw opacity=1 ][fill={rgb, 255:red, 255; green, 255; blue, 255 }  ,fill opacity=1 ] (119,144.8) -- (156.07,144.8) -- (156.07,176) -- (119,176) -- cycle ;
		\draw  [color={rgb, 255:red, 255; green, 255; blue, 255 }  ,draw opacity=1 ][fill={rgb, 255:red, 255; green, 255; blue, 255 }  ,fill opacity=1 ] (246.5,138.8) -- (283.57,138.8) -- (283.57,170) -- (246.5,170) -- cycle ;
		\draw (263.91,199.4) node [anchor=north] [inner sep=0.75pt]  [font=\normalsize] [align=left] {Observe};
		\draw (141.36,199.4) node [anchor=north] [inner sep=0.75pt]  [font=\normalsize] [align=left] {Model};
		\draw (202.94,215.33) node [anchor=north] [inner sep=0.75pt]  [font=\footnotesize] [align=left] {Inference};
		\draw (138.03,146.89) node [anchor=north] [inner sep=0.75pt]  [font=\footnotesize] [align=left] {Optimize};
		\draw (138.03,161.89) node [anchor=north] [inner sep=0.75pt]  [font=\footnotesize] [align=left] {EIG};
		\draw (266.03,141.89) node [anchor=north] [inner sep=0.75pt]  [font=\footnotesize] [align=left] {Run};
		\draw (266.03,156.89) node [anchor=north] [inner sep=0.75pt]  [font=\footnotesize] [align=left] {Experiment};
		\draw (203.91,122.4) node [anchor=north] [inner sep=0.75pt]  [font=\normalsize] [align=left] {Design};
	\end{tikzpicture}
	\caption{Traditional BAD iterates between choosing designs by maximizing the incremental EIG, running the experiment, and updating our beliefs via Bayesian inference.}
	\label{fig:boed_approaches_traditional}
\end{figure}
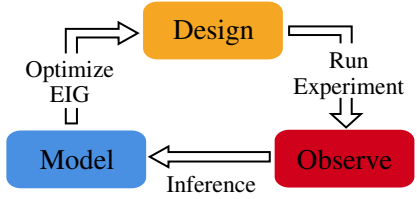

%% file: figures/dad.tex
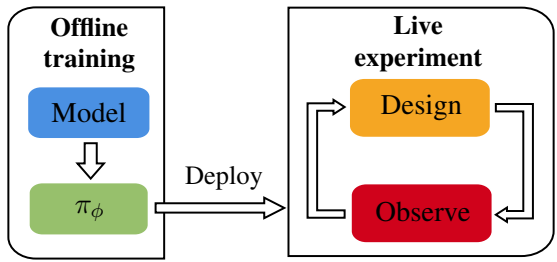
\begin{figure}[t]
	\tikzset{every picture/.style={line width=0.75pt}}
	\begin{tikzpicture}[x=0.75pt,y=0.75pt,yscale=-0.85,xscale=1.1]
		\draw   (532.17,216.83) -- (515.17,216.83) .. controls (515.17,216.83) and (515.17,216.83) .. (515.17,216.83) -- (515.17,148.88) .. controls (515.17,148.88) and (515.17,148.88) .. (515.17,148.88) -- (527.07,148.88) -- (527.07,146.83) -- (532.17,151.08) -- (527.07,155.33) -- (527.07,153.29) -- (519.57,153.29) .. controls (519.57,153.29) and (519.57,153.29) .. (519.57,153.29) -- (519.57,212.43) .. controls (519.57,212.43) and (519.57,212.43) .. (519.57,212.43) -- (532.17,212.43) -- cycle ;
		\draw   (601.5,149.29) -- (618.5,149.29) .. controls (618.5,149.29) and (618.5,149.29) .. (618.5,149.29) -- (618.5,217.24) .. controls (618.5,217.24) and (618.5,217.24) .. (618.5,217.24) -- (606.6,217.24) -- (606.6,219.29) -- (601.5,215.04) -- (606.6,210.79) -- (606.6,212.83) -- (614.09,212.83) .. controls (614.09,212.83) and (614.09,212.83) .. (614.09,212.83) -- (614.09,153.7) .. controls (614.09,153.7) and (614.09,153.7) .. (614.09,153.7) -- (601.5,153.7) -- cycle ;
		\draw  [color={rgb, 255:red, 255; green, 255; blue, 255 }  ,draw opacity=1 ][fill={rgb, 255:red, 151; green, 192; blue, 108 }  ,fill opacity=0.58 ] (387.33,202.13) .. controls (387.33,198.65) and (390.15,195.83) .. (393.63,195.83) -- (436.03,195.83) .. controls (439.51,195.83) and (442.33,198.65) .. (442.33,202.13) -- (442.33,221.03) .. controls (442.33,224.51) and (439.51,227.33) .. (436.03,227.33) -- (393.63,227.33) .. controls (390.15,227.33) and (387.33,224.51) .. (387.33,221.03) -- cycle ;
		\draw   (613.23,92.07) .. controls (621.39,92.07) and (628,98.68) .. (628,106.84) -- (628,224.81) .. controls (628,232.97) and (621.39,239.58) .. (613.23,239.58) -- (506.43,239.58) .. controls (506.43,239.58) and (506.43,239.58) .. (506.43,239.58) -- (506.43,92.07) .. controls (506.43,92.07) and (506.43,92.07) .. (506.43,92.07) -- cycle ;
		\draw   (390.62,241.08) .. controls (383.65,241.08) and (378,235.43) .. (378,228.46) -- (378,104.53) .. controls (378,97.56) and (383.65,91.91) .. (390.62,91.91) -- (449.43,91.91) .. controls (449.43,91.91) and (449.43,91.91) .. (449.43,91.91) -- (449.43,241.08) .. controls (449.43,241.08) and (449.43,241.08) .. (449.43,241.08) -- cycle ;
		\draw  [fill={rgb, 255:red, 255; green, 255; blue, 255 }  ,fill opacity=1 ] (445.5,208.23) -- (495.31,208.23) -- (495.31,204.93) -- (504.5,211.01) -- (495.31,217.08) -- (495.31,213.78) -- (445.5,213.78) -- cycle ;
		\draw  [color={rgb, 255:red, 255; green, 255; blue, 255 }  ,draw opacity=1 ][fill={rgb, 255:red, 74; green, 144; blue, 226 }  ,fill opacity=0.35 ] (386.63,145.08) .. controls (386.63,141.63) and (389.43,138.83) .. (392.88,138.83) -- (436.88,138.83) .. controls (440.34,138.83) and (443.13,141.63) .. (443.13,145.08) -- (443.13,163.83) .. controls (443.13,167.28) and (440.34,170.08) .. (436.88,170.08) -- (392.88,170.08) .. controls (389.43,170.08) and (386.63,167.28) .. (386.63,163.83) -- cycle ;
		\draw  [fill={rgb, 255:red, 255; green, 255; blue, 255 }  ,fill opacity=1 ] (418.52,172.58) -- (418.52,184.58) -- (421.82,184.58) -- (415.75,193.58) -- (409.68,184.58) -- (412.98,184.58) -- (412.98,172.58) -- cycle ;
		\draw  [color={rgb, 255:red, 255; green, 255; blue, 255 }  ,draw opacity=1 ][fill={rgb, 255:red, 208; green, 2; blue, 27 }  ,fill opacity=0.38 ] (535,203.53) .. controls (535,199.61) and (538.18,196.43) .. (542.1,196.43) -- (592.9,196.43) .. controls (596.82,196.43) and (600,199.61) .. (600,203.53) -- (600,224.83) .. controls (600,228.75) and (596.82,231.93) .. (592.9,231.93) -- (542.1,231.93) .. controls (538.18,231.93) and (535,228.75) .. (535,224.83) -- cycle ;
		\draw  [color={rgb, 255:red, 255; green, 255; blue, 255 }  ,draw opacity=1 ][fill={rgb, 255:red, 245; green, 166; blue, 35 }  ,fill opacity=0.53 ] (534.17,140.93) .. controls (534.17,137.01) and (537.35,133.83) .. (541.27,133.83) -- (592.07,133.83) .. controls (595.99,133.83) and (599.17,137.01) .. (599.17,140.93) -- (599.17,162.23) .. controls (599.17,166.15) and (595.99,169.33) .. (592.07,169.33) -- (541.27,169.33) .. controls (537.35,169.33) and (534.17,166.15) .. (534.17,162.23) -- cycle ;
		\draw (567.35,151.6) node  [font=\normalsize] [align=left] {Design};
		\draw (568,213.1) node  [font=\normalsize] [align=left] {Observe};
		\draw (415.33,212.5) node  [font=\normalsize] [align=left] {$\pi_{\phi}$};
		\draw (415,114.17) node  [font=\small,color={rgb, 255:red, 0; green, 0; blue, 0 }  ,opacity=1 ] [align=left] {\begin{minipage}[lt]{38.94pt}\setlength\topsep{0pt}
				\begin{center}
					\textbf{Offline }\\\textbf{training}
				\end{center}
			\end{minipage}};
		\draw (476.79,193.17) node  [font=\normalsize,color={rgb, 255:red, 0; green, 0; blue, 0 }  ,opacity=1 ] [align=left] {\textcolor[rgb]{0,0,0}{{\small Deploy}}};
		\draw (565.82,113.33) node  [font=\small,color={rgb, 255:red, 0; green, 0; blue, 0 }  ,opacity=1 ] [align=left] {\begin{minipage}[lt]{51.7pt}\setlength\topsep{0pt}
				\begin{center}
					\textbf{Live }\\\textbf{experiment}
				\end{center}
			\end{minipage}};
		\draw (414.88,154.46) node  [font=\normalsize] [align=left] {Model};
	\end{tikzpicture}
	\caption{DAD: a design policy network, $\pi_{\phi}$, is learnt upfront, then deployed near-instantaneously during the experiment.}
	\label{fig:boed_approaches_idad}
\end{figure}